# On MAP inference by MWSS on Perfect Graphs

**Adrian Weller**
Department of Computer Science
Columbia University
New York NY 10027
adrian@cs.columbia.edu

**Tony Jebara**
Department of Computer Science
Columbia University
New York NY 10027
jebara@cs.columbia.edu

**Abstract**

Finding the most likely (MAP) configuration of a Markov random field (MRF) is NP-hard in general. A promising, recent technique is to reduce the problem to finding a maximum weight stable set (MWSS) on a derived weighted graph, which if perfect, allows inference in polynomial time. We derive new results for this approach, including a general decomposition theorem for MRFs of any order and number of labels, extensions of results for binary pairwise models with submodular cost functions to higher order, and an exact characterization of which binary pairwise MRFs can be efficiently solved with this method. This defines the power of the approach on this class of models, improves our toolbox and expands the range of tractable models.

## 1 INTRODUCTION

Markov random fields (MRFs), also termed undirected probabilistic graphical models, are a central tool in machine learning with wide use in many areas including speech recognition, vision and computational biology. A model $(V, \Psi)$ is specified by a set of $n$ variables $V = \{X_1, \ldots, X_n\}$ together with (log) potential functions over subsets $c$ of $V$, $\Psi = \{\psi_c : c \in C \subseteq \mathcal{P}(V)\}$, where $\mathcal{P}(V)$ is the powerset of $V$. In this paper, each variable $X_i$ may take finite $k_i$ possible values which we label $\{0, \ldots, k_i - 1\}$. Write $x = (x_1, \ldots, x_n)$ for one particular complete configuration and $x_c$ for a configuration just of the variables in $c$. A potential function $\psi_c$ maps each possible setting $x_c$ of its variables $c$ to a real number $\psi_c(x_c)$.

Identifying a configuration of variables that is most likely, termed *maximum a posteriori* or MAP inference, is very useful in many contexts, yet in general is NP-hard (Shimony, 1994). In our notation this is the combinatorial problem of identifying[1]

$$x^* = \arg\max_{x=(x_1,\ldots,x_n)} \sum_{c \in C} \psi_c(x_c). \quad (1)$$

In general, an MRF may be considered a hypergraph together with associated $\psi_c$ functions (see section 2.1 for definitions). A popular alternative representation is a factor graph, which is a bipartite graph where the variables $V$ form one stable partition and each $c \in C$ is a node in the other partition, with an edge from $c$ to each variable it contains. In the special case that all variables $X_i$ take values only in $\mathbb{B} = \{0, 1\}$, the model is said to be *binary*. If $|c| \leq 2 \; \forall c \in C$ then the model is *pairwise*. Binary pairwise models play a key role in computer vision both directly and as critical subroutines in solving more complex problems (Pletscher & Kohli, 2012). Note that it is possible to convert a general MRF into an equivalent binary pairwise model (Yedidia et al., 2001; Ravikumar & Lafferty, 2006), though this may lead to a much larger state space.

### 1.1 RELATED WORK

It is well-known that the MAP estimate can be recovered for junction trees and acyclic graphical models using dynamic programming, junction tree algorithms, as well as max-product message passing (Bertelé & Brioschi, 1972; Pearl, 1988; Wainwright & Jordan, 2008). Such approaches hinge on the graph having bounded dimension or low tree-width, which is indeed the case for many useful Bayesian networks. Subsequently, graphical models with more general (and often dense) topologies yet whose potentials are constrained to be binary pairwise associative (ferromagnetic) functions were shown to be solvable efficiently using graph-cuts or network flow (Greig et al., 1989; Boykov & Kolmogorov, 2004). Many computer vision and image processing problems can be handled by this class of models. More recently, MAP estimation for cyclic graphical models involving matching and *b*-matching problems[2] was shown to be solvable efficiently using the max-product algorithm (Bayati et al., 2005; Huang & Jebara, 2007; Sanghavi et al., 2008; Bayati et al., 2008). In previous work, these three known cases were all shown to compile to a maximum weight stable set problem on a perfect graph, which is known to be solvable in polynomial time (Jebara, 2009; Jebara, 2012). This paper derives new results for this approach, first described in (Jebara, 2009; Sanghavi et al., 2009), and examines which other models may be handled in this manner.

---

[1]This formulation assumes each configuration has probability $> 0$. When this is not the case, typically 0 may be replaced by a sufficiently small $\epsilon$. Also *cost* functions are the negative of our $\psi$s, thus submodular cost functions are equivalent to supermodular $\psi$s.

An earlier method examining triangulated[3] microstructure graphs was presented (Jégou, 1993) in the context of constraint satisfaction problems (CSPs). Valued CSPs (VCSPs) use soft constraints with explicit costs, and are closely related to MAP inference problems. Many other techniques have been developed, including optimal soft arc consistency (Cooper et al., 2010), belief propagation (Weiss et al., 2007) and linear program relaxations (Sontag et al., 2008), which may be considered to proceed through identifying helpful reparameterizations (see section 2.5).

### 1.2 CONTRIBUTION AND SUMMARY

In section 2, we present important preliminary terms and results from graph theory, and on the approach of MAP inference via MWSS on a derived graph called an NMRF (a *nand Markov random field*, see section 2.4). This reviews previous work and introduces some novel concepts needed later.

In section 3 we derive a general decomposition theorem for mapping MRFs to NMRFs, which can be used to break apart a complex problem into smaller parts that overlap only on single variables. In situations where there are only a few of these overlapping variables, one could solve each subproblem and use a brute force enumeration approach over all combinations of the overlapping variables to find the global optimum, but this is clearly exponential in the number of overlapping variables. Our approach, in contrast, runs in polynomial time even for $\Omega(n)$ overlapping variables. This general result applies for potential functions $\psi_c$ of any order, and variables with any number of labels. Note that each subproblem could have high treewidth.

In section 4 we apply this general result specifically to pairwise models, focusing on the binary case to derive features of corresponding NMRFs. Applying these results, we proceed in section 5 to build towards Theorem 19, which provides a precise characterization of which binary pairwise MRFs map to perfect NMRFs for all valid $\psi_c$, and hence are amenable to this approach for efficient MAP inference.

In section 6 we explore a different direction, generalizing the previous result (Jebara, 2012) that binary pairwise MRFs with submodular cost functions can always be mapped to bipartite NMRFs (a special case of perfect graphs, admitting faster inference). We show that a bipartite pruned NMRF is obtained, for any topology, for third order cost functions iff they are submodular, and demonstrate that for interactions of order $\geq 4$, submodularity is a necessary but strictly insufficient condition. Section 7 provides a conclusion and outlines future work.

## 2 BACKGROUND

### 2.1 TERMS FROM GRAPH THEORY

We follow standard definitions and omit some familiar terms, see (Diestel, 2010). A *graph* $G(V, E)$ is a set of vertices $V$, and edges $E \subseteq V \times V$. Let $n = |V|$ and $m = |E|$. Throughout this paper, unless otherwise specified, all graphs are finite and *simple*, that is a vertex may not be adjacent to itself (no loops) and each pair of vertices may have at most one edge (no multiple edges).

The *complete* graph on $n$ vertices, written $K_n$, has all $\binom{n}{2}$ edges. A *path* of length $n$ is a graph $P_n$ with $n$ edges connecting $n+1$ vertices as $v_1 - v_2 - \cdots - v_n - v_{n+1}$. An *induced subgraph* $H(U, F)$ of a graph $G(V, E)$ is a graph on some subset of the vertices $U \subseteq V$, inheriting all edges with both ends in $U$, so $F = \{(v, w) \in E : v, w \in U\}$. The union of two subgraphs, $H_1(V_1, E_1)$ and $H_2(V_2, E_2)$ of a graph $G(V, E)$, written $H_1 + H_2$, is the induced subgraph of $G$ on $V_1 \cup V_2$.

A *hypergraph* $(V, E)$ is a generalization of a graph where the elements of $E$ are any non-empty subsets of $V$, not necessarily of size two. A general MRF may be regarded as a hypergraph $(V, C)$ together with functions $\{\psi_c\} \, \forall c \in C$. For the special case of a pairwise model, the structural relationships are naturally interpreted as a graph.

A graph is *connected* if there is a path connecting any two vertices. A *cut vertex* of a connected graph $G$ is a vertex $v \in V$ such that deleting $v$ disconnects $G$. A graph is *2-connected*, equivalently *biconnected*, if it is connected and contains no cut vertex. A *block* is a maximal connected subgraph with no cut vertex of the subgraph. Every block is either $K_2$ (two vertices joined by an edge) or a maximal 2-connected subgraph containing a cycle. Different blocks of $G$ overlap on at most one vertex, which must be a cut vertex. Hence $G$ can be written as the union of its blocks with every edge in exactly one block. These blocks are connected without cycles in the *block tree* for $G$.

---
[2]These graphical models involve topological constraints as well as various constraints on the potential functions (not simply associativity or submodularity).

[3]Triangulated, or chordless, graphs are a subclass of perfect graphs.

A *stable* set in a graph is a set of vertices, no two of which are adjacent. A *weighted graph* $(V, E, w)$ is a graph with a nonnegative real value for each vertex, called its *weight* $w(v)$. A *maximum weight stable set (MWSS)* is a stable set with maximum possible weight. A *maximal maximum weight stable set (MMWSS)* is a MWSS of maximal cardinality (this is useful in our context since, after reparameterization, we may have many nodes with 0 weight, see sections 2.4 and 2.5).

A *clique* in a graph is a set of vertices, of which every pair is adjacent. The *clique number* of a graph $G$, written $\omega(G)$, is the maximum size of a clique in $G$.

The *complement* of a graph $G(V, E)$ is the graph $\bar{G}(V, F)$ on the same vertices with an edge in $F$ iff it is not in $E$. Hence a clique is the complement of a stable set and vice versa.

A *coloring* of a graph is a map from its vertices to the integers (considered the colors of the vertices) such that no two adjacent vertices share the same color. The *chromatic number* of a graph $G$, written $\chi(G)$, is the minimum number of colors required to color it. Observe that clearly $\chi(G) \geq \omega(G)$ for any graph $G$.

A graph $G$ is *perfect* iff $\chi(H) = \omega(H)$ for all induced subgraphs $H$ of $G$. As examples, any bipartite or chordal graph is perfect. Related concepts (see Theorem 5) are: a *hole* in a graph $G$ is an induced subgraph which is a cycle of length $\geq 4$ (note this means the cycle must be chordless); an *antihole* is an induced subgraph whose complement is a hole. A hole or antihole is *odd* if it has an odd number of vertices. Note that, as a special case, a hole with 5 vertices is equivalent to an antihole of the same size. It is easily shown that odd holes and antiholes are not perfect.

## 2.2 FURTHER TERMS

This section may be skipped on a first reading, and referred to later for definitions.

A *clique group* for a set of variables $c$ is a clique in an NMRF corresponding to all possible settings $x_c$ of those variables of its MRF, see section 2.4.

An *snode* is a node in an NMRF relating to a setting of a single variable from its MRF. Equivalently, it is a node from a clique group deriving from $c = \{X_i\}$ for some $i$. An *enode* is a node from a clique group deriving from some $c \in C$ with $|c| \geq 2$. For example, when considering binary pairwise models, an enode derives from an edge of the MRF.

For a graph $(V, E)$, if $X \subseteq V$ and $v \in V \setminus X$ then $v$ is *complete* to $X$ if $v$ is adjacent to every member of $X$. If $X, Y \subseteq V$ are disjoint, then $X$ is *complete* to $Y$ if every vertex in $X$ is complete to $Y$.

A *cutset* $S$ of a graph $G$ is a set of vertices $S \subseteq V(G)$ s.t. $G \setminus S$ is disconnected. A *star-cutset* $S$ of $G$ is a cutset s.t. $\exists$ some $x \in S$ s.t. $x$ is complete to $S \setminus \{x\}$.

A *signed graph* (Harary, 1953) is a graph $(V, E)$ together with one of two possible signs for each edge. This is a natural structure when considering binary pairwise models, where we characterize edges as either *associative* or *repulsive*, see section 2.6. When discussing signed graphs, we use the notation $\oplus$ to show an associative edge, and $\ominus$ for a repulsive edge. For example, $x \oplus y \ominus z$ is a graph with 3 vertices $x$, $y$ and $z$, and two edges, where $x$ and $y$ are adjacent via an associative edge, and $y$ and $z$ are adjacent via a repulsive edge.

A *frustrated cycle* in a signed graph is a cycle with an odd number of repulsive edges.

A $B_R$ structure (see Figure 1 for an example) is a signed graph over variables $V$ with associative edges $E_A$ and repulsive edges $E_R$ s.t. $(V, E_R)$ is bipartite and $\exists$ a disjoint bipartition $V = V_1 \cup V_2$ with all $E_R$ crossing between the partitions $V_1 - V_2$, and no $E_A$ crossing between them. Either $E_A$ or $E_R$ may be empty, so for example, this includes any signed graph with only associative edges.

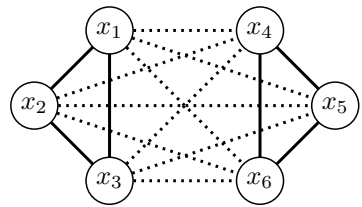

Figure 1: A $B_R$ structure. Solid (dashed) edges are associative (repulsive). Deleting any edges maintains the $B_R$ property.

A $T_{m,n}$ structure (see Figure 2 for an example) is a 2-connected signed graph containing $m + n \geq 1$ triangles on a common base given by: 2 base vertices $s, t$ connected via a repulsive edge, so $s \ominus t$; together with $m \geq 0$ vertices $r_i$, each adjacent only to $s$ and $t$ via repulsive edges, so $s \ominus r_i \ominus t$; and $n \geq 0$ vertices $a_i$, each adjacent only to $s$ and $t$ via associative edges, so $s \oplus a_i \oplus t$. Note $T_{m,n}$ would be bipartite, with $\{s, t\}$ as one partition and all other vertices in the other, except that we have the repulsive edge $s \ominus t$.

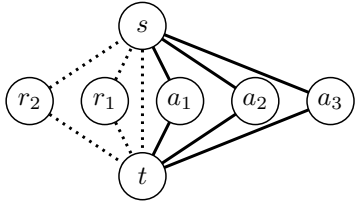

Figure 2: A $T_{m,n}$ structure with $m = 2$ and $n = 3$. Solid (dashed) edges are associative (repulsive).

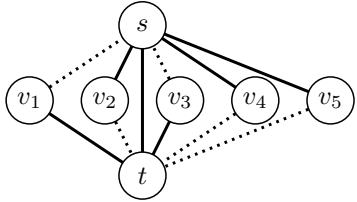

Figure 3: A $U_n$ structure with $n = 5$. Solid (dashed) edges are associative (repulsive).

A $U_n$ structure (see Figure 3 for an example) is a 2-connected signed graph containing $n \geq 1$ triangles on a common base given by: 2 base vertices $s, t$ connected via an associative edge, so $s \oplus t$; together with $n \geq 1$ vertices $v_i$, each adjacent only to $s$ and $t$ via one associative and one repulsive edge (either way), so either $s \oplus v_i \ominus t$ or $s \ominus v_i \oplus t$.

Note that $U_1$ is the same as $T_{0,1}$ but this is the only overlap. In Lemma 18, we show that $T_{m,n}$ and $U_n$ structures are the only 2-connected signed graphs containing a frustrated cycle that map to a perfect NMRF.

## 2.3 PROPERTIES OF PERFECT GRAPHS

### 2.3.1 Complexity of MWSS

Our approach to MAP inference is to reduce the problem to finding a maximum weight stable set on a derived weighted graph, as described in section 2.4. This is helpful only if we can find a MWSS efficiently, yet in general this is still an NP-hard problem for a graph with $N$ vertices. However, if the derived graph is perfect[4], then a MWSS may be found in polynomial time via the ellipsoid method (Grötschel et al., 1984).

Faster exact methods (Yildirim & Fan-Orzechowski, 2006) based on semidefinite programming are possible in $O(N^6)$ and are improved using primal-dual methods (Chan et al., 2009). Alternatively, linear programming can solve MWSS problems but requires $O(N^3 \sqrt{n_K})$ time where $n_K$ is the number of maximal cliques in the graph (Jebara, 2009; Jebara, 2012). Clearly, whenever $n_K$ is small, linear programming can be more ef-ficient than semidefinite programming. However, in the worst case, $n_K$ may be exponentially large in $N$ which makes linear programming useful only in some cases. Message-passing methods can also be applied for finding the maximum weight stable set in a perfect graph though they too become inefficient for graphs with many cliques (Foulds et al., 2011; Jebara, 2012).

---

[4]There are a few other classes of graphs that also admit efficient MWSS, such as *claw-free* graphs, where significant recent advances have been made (Faenza et al., 2011), but so far these have not been useful in analyzing MRFs.

Where other methods exist for solving exact MAP inference, the reduction to MWSS is typically not the fastest method, yet there is hope for improvement since the field is advancing rapidly, with significant breakthroughs in recent years (Chudnovsky et al., 2006; Faenza et al., 2011).

### 2.3.2 Other properties

There is a rich literature on perfect graphs. We highlight key results used later in this paper.

**Theorem 1** ((Gallai, 1962)). *The graph obtained by pasting two perfect graphs on a clique is perfect.*

**Theorem 2** ((Chvátal, 1985)). *The graph obtained by pasting two perfect graphs on a star-cutset is perfect.*

**Theorem 3** (Substitution Lemma, (Lovász, 1972)). *The graph obtained by substituting one perfect graph for a vertex of another perfect graph is also perfect.*

Here, substituting $H$ for $x$ in $G$ means deleting $x$ and joining every vertex of $H$ to those vertices of $G$ which were adjacent to $x$.

**Theorem 4** (Weak Perfect Graph Theorem, (Lovász, 1972)). *A graph is perfect iff its complement is perfect.*

**Theorem 5** (Strong Perfect Graph Theorem 'SPGT' (Chudnovsky et al., 2006)). *A graph is perfect iff it contains no odd hole or antihole.*

## 2.4 MAP REDUCTION TO MWSS

Given an MRF model $(V, \Psi)$, construct a *nand Markov random field (NMRF)*, see (Jebara, 2009):

- A weighted graph $N(V_N, E_N, w)$ with vertices $V_N$, edges $E_N$ and a weight function $w : V_N \to \mathbb{R}_{\geq 0}$.
- Each $c \in C$ of the original model maps to a *clique group* of $N$ which contains one node for each possible configuration $x_c$, all pairwise adjacent.
- Generally, nodes are adjacent iff they have inconsistent settings for any variable $X_i$.
- Nonnegative weights of each node in $N$ are set as $\psi_c(x_c) - \min_{x_c} \psi_c(x_c)$, see section 2.5 for an explanation of the subtraction.

See Figure 4 for an example. (Jebara, 2012) proved that a maximal cardinality set of consistent configuration nodes in $N$ with greatest total weight, i.e. a MMWSS of $N$ (see section 2.1), will identify a globally consistent configuration of all variables of the original MRF that solves the MAP inference problem (1).

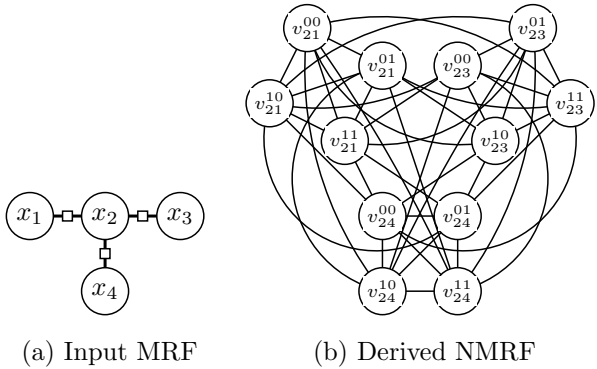

(a) Input MRF  (b) Derived NMRF

Figure 4: An example of mapping an MRF with binary variables (shown as a factor graph) to an NMRF (subscripts denote the factor variables $c$ and superscripts denote the configuration $x_c$).

*Sketch proof:* (Slightly different to (Jebara, 2012), this will allow us to extend the result after discussing pruning in section 2.5.) A MMWSS $S$ is consistent by construction and clearly contains at most one node from each clique group. It remains to show it has at least one from each clique group. Suppose a clique group has no representative. Identify a member of this group which could be added to $S$, establishing a contradiction since $S$ is maximal, as follows: the group overlaps with some variables of $S$, copy the settings of these; for all other variables in the group, pick any setting. Note that if we do not insist on a maximal MWSS, it is possible that we do not get a representative for some clique groups and hence do not obtain a complete MAP configuration for the initial MRF.

## 2.5 REPARAMETERIZATIONS AND PRUNING

A *reparameterization* is a transformation

$$\{\psi_c\} \to \{\psi'_c\} \text{ s.t. } \forall x, \sum_{c\in C}\psi_c(x_c) = \sum_{c\in C}\psi'_c(x_c)+\text{constant}.$$

This clearly does not modify (1) but can be helpful to simplify the problem.

One particular reparameterization is to add a constant just to any $\psi_c$ function, since any consistent configuration has exactly one setting for each group of variables $c$. Hence we may subtract the minimum $\psi_c(x_c)$ and assume that in each clique group of $N$, the minimum weight of a node is exactly zero. The earlier reduction result in section 2.4 holds provided we insist on a *maximal* MWSS. To find a MMWSS, it is sufficient first to remove or *prune* the zero weight nodes, find a MWSS on the remaining graph, then reintroduce a maximal number of the zero weight nodes while maintaining stability of the set. Different reparameterizations will yield different pruned NMRFs. By the earlier argument: MWSS will find one member from each of some of the clique groups, then we can always find one of the zero weight nodes to add from each of the remaining groups using any greedy method. Hence we have shown the following result.

**Lemma 6.** *MAP inference on an MRF is tractable provided $\exists$ an efficient reparameterization s.t. the MRF maps to a perfect pruned NMRF.*

## 2.6 SINGLETON TRANSFORMATIONS, BINARY PAIRWISE MRFS AND ASSOCIATIVITY

Another useful reparameterization is what we term a *singleton transformation*, which is a change in one or more $\psi$ functions for a single variable, with corresponding changes to a higher order term which brings it to a convenient form.

Considering binary pairwise models only, it is easily shown that a reparameterization of an edge via singleton transformations, $\begin{pmatrix}\psi_{00} & \psi_{01}\\ \psi_{10} & \psi_{11}\end{pmatrix} \to \begin{pmatrix}\psi'_{00} & \psi'_{01}\\ \psi'_{10} & \psi'_{11}\end{pmatrix}$ is valid iff $\psi_{00}+\psi_{11}-\psi_{01}-\psi_{10} = \psi'_{00}+\psi'_{11}-\psi'_{01}-\psi'_{10}$. Hence this quantity, which we call the *associativity* of the edge, is invariant with respect to any singleton transformation, and thus is well defined.

We describe an edge as either *associative*[5], in which case it tends to pull its two end vertices toward the same value, or *repulsive*, in which case it tends to push its two end vertices apart to different values, according to whether its associativity is $>0$ or $<0$. An edge with 0 associativity may be removed since we may transform its edge potential to the zero matrix. A binary pairwise model is associative iff every one of its edges is associative.

An associative edge may be reparameterized s.t. three of its entries are 0, and therefore may be pruned, leaving only either $\psi'_{00}$ or $\psi'_{11}$ (or both, though for our purposes of mapping to a perfect NMRF, it is always easier to prune more nodes) with a positive value. Similarly, we may reparameterize a repulsive edge $x \ominus y$ to leave only a $(x=0, y=1)$ or $(x=1, y=0)$ node.[6]

---

[5]Other equivalent terms used are *attractive*, *ferromagnetic* or *regular*. This is equivalent to $\psi$ for the edge being supermodular, or having submodular cost function.

[6]For repulsive edges, selecting one or other form is exactly analogous to choosing an *orientation* of the edge, $x \to y$ or $x \leftarrow y$. Further, such enodes from repulsive edges are adjacent iff their directed edges connect 'head to tail', hence the induced subgraph of an NMRF on these

## 2.7 SINGLETON CLIQUE GROUPS

Since typically we would like to allow any finite values for singleton potential functions, and singleton transformations as described in section 2.6 without restriction, in this paper we assume that any NMRF includes the complete clique group for each of the single variables of its MRF. In particular contexts, however, one may drop this requirement, and since this would remove nodes from the NMRF, it can only help to show perfection (any induced subgraph of a perfect graph is perfect), though then sometimes care must be taken to confirm the decomposition result of Theorem 7.

## 3 NEW RESULTS FOR ALL MRFS

**Theorem 7** (MRF Decomposition)**.** *If $MRF_A(V_A, \Psi_A)$ and $MRF_B(V_B, \Psi_B)$ both map to perfect NMRFs $N_A$ and $N_B$, and have exactly one variable s in common, i.e. $V_A \cap V_B = \{s\}$, with consistent $\psi_s$, then the combined $MRF'(V_A \cup V_B, \Psi_A \cup \Psi_B)$ maps to an NMRF $N'$ which is also perfect. The converse is true by the definition of perfect graphs.*

*Proof.* [7] See section 2.2 for notation. We may assume both $\Psi_A$ and $\Psi_B$ contain the same $\psi_s$ forming the complete $s$ clique group $K_s$ in $N_A$ and $N_B$ (see section 2.7, though in fact this Theorem holds more generally, provided only that both $N_A$ and $N_B$ have the same nodes from the clique group for $s$).

Let the possible values of $s$ be $\{0, \ldots, k-1\}$, and $s_i$ be the snode corresponding to $(s = i)$. Let $A_i$ be all those vertices of $N_A \setminus \{s_i\}$ which have setting $s = i$, similarly define $B_i$ for $N_B$. Observe that $A_i$ is complete to $A_j$ for all $i \neq j$, and similarly for $B_i$. $N'$ is the result of pasting $N_A$ and $N_B$ on $K_s$, together with all edges from $A_i$ to $B_j$ if $i \neq j$.

Hence $N'$ admits a star-cutset given by $X = K_s + A_1 + \cdots + A_k + B_1 + \cdots + B_k$ with $s_0$ complete to $X \setminus \{s_0\}$. Thus by Theorem 2, it is sufficient to show that $N_A + X$ and $N_B + X$ are each perfect. But this is true by Theorem 3, since $N_A + X = N_A + B_1 + \cdots + B_k$ may be obtained from $N_A$ by substituting (via Theorem 3) $B_i + s_i$ for $s_i$, $i = 1, \ldots, k$; and similarly for $N_B$. □

### 3.1 BLOCK DECOMPOSITION

Theorem 7 is a powerful tool for analyzing MRFs of any order and number of labels. As a special case, we have an immediate corollary.

---

repulsive enodes is exactly a directed line graph of $(V, E_R)$.

[7]This proof, due to Maria Chudnovsky, is shorter and neater than the authors' original.

**Theorem 8.** *A pairwise MRF maps to a perfect NMRF for all valid $\psi$ iff each of its blocks maps to a perfect NMRF.*

This provides an elegant way to derive a previous result (Jebara, 2009):

**Theorem 9.** *A pairwise MRF whose graph structure is a tree (i.e. no cycles) maps to a perfect NMRF.*

*Proof.* By Theorem 8, we need only consider one edge together with its two end vertices (then use induction). The edge clique group together with each one of the singleton clique groups is the complement of a bipartite graph, hence is perfect (by Theorem 4). Now paste the two together on the edge clique group to show the whole is perfect (by Theorem 1). □

We show the following further general result.

**Lemma 10.** *Neither an odd hole H nor an odd antihole A in a NMRF can contain $\geq 2$ members, say $s_1$ and $s_2$, of any singleton clique group.*

*Proof.* In $H$, $s_1$ and $s_2$ must be next to each other, then moving out round $H$ one node in each direction, we cannot avoid a chord, contradiction. In $A$, there must be at least 2 nodes between $s_1$ and $s_2$ in at least one direction. Taking this way round $A$, the node next to $s_1$ must be adjacent to $s_2$ but not $s_1$, so has setting $s = 1$. Continuing round $A$, the next node must be adjacent to $s_1$, so must have an $s$ value $\neq 1$ but then it is adjacent to its predecessor, contradiction. □

## 4 NEW RESULTS FOR BINARY PAIRWISE MRFS

**Lemma 11.** *Let M be a binary pairwise MRF. $\exists$ a reparameterization s.t. M maps to perfect pruned NMRF $\Leftrightarrow$ $\exists$ a reparameterization with just one enode per edge in the pruned NMRF which is perfect.*

*Proof.* ($\Leftarrow$) is clear. ($\Rightarrow$) see section 2.6. With a standard reparameterization, we may always achieve just one pruned enode (either 00 or 11 for associative, 01 or 10 for repulsive) from those already present. The result follows from the definition of a perfect graph. □

Therefore henceforth, when referring to a pruned NMRF of a binary pairwise MRF, we may assume just one enode per edge.

**Lemma 12.** *An antihole A of size $\geq 7$ can never occur in a pruned NMRF N from a binary pairwise MRF M.*

*Proof.* Suppose $A$ exists containing an snode, WLOG say $s_0$. This must be adjacent to $\geq 4$ nodes in $A$, all of which must have $s = 1$ settings. The 2 closest to $s_0$ around $A$ one way must both be adjacent to the closest to $s_0$ around $A$ the other way, which cannot be achieved, hence $A$ must contain only enodes. By Lemma 11, we have only one enode per edge of $M$. Two enodes are adjacent in $N$ if they have one end in common with different settings - since only 2 settings are possible, a triangle in $N$ must derive from edges in $M$ that formed a triangle. Given 2 enodes which are adjacent, there is exactly one possible third enode with which they can form a triangle (e.g. for $s0t1$ and $t0u0$, $s1u1$ is the unique third possible enode). Yet $A$ must contain $\geq 2$ triangles which have the same 2 members but a different third member, contradiction. □

Since an antihole of size 5 is equivalent to a hole of the same size, SPGT (Theorem 5) gives the following.

**Lemma 13.** *For a binary pairwise MRF, a pruned NMRF is perfect $\Leftrightarrow$ it contains no odd hole.*

## 5 WHICH BINARY PAIRWISE MRFS YIELD PERFECT NMRFS

By Theorem 8, we need only consider 2-connected graphs $G$ (considering both associative and repulsive edges), and by Lemma 13 we need only check for odd holes. $G$ either contains a frustrated cycle or does not. If it does, we shall see that $G$ must have the form $T_{m,n}$ or $U_n$. If not, we show $G$ must have the form $B_R$. See section 2.2 for definitions.

**Lemma 14** ((Harary, 1953)). *The following are equivalent properties for a signed graph $G$ on vertices $V$:*

1. *$G$ contains no frustrated cycle*
2. *$G$ is of the form $B_R$*
3. *$G$ is flippable to fully associative*

(1)⇔(2) by a variant of the standard proof that a graph is bipartite iff it has no odd cycle, considering repulsive edges. (3) means $\exists$ some subset $S \subseteq V$ s.t. if we replace each $X_i \in S$ by $Y_i = 1 - X_i$, and modify potential functions accordingly, thereby flipping the nature of each edge incident to $X_i$ between associative and repulsive, then all edges of $G$ can be made associative. (2)⇔(3) by setting $S$ as either partition.

**Theorem 15.** *A binary pairwise MRF with the form $B_R$ maps efficiently to a bipartite NMRF $N$.*

*Proof.* Let the partitions of the variables be $S$ and $T$ with snodes $\{s_i^0, s_i^1\}$ from $S$, and $\{t_j^0, t_j^1\}$ from $T$. Choose a reparameterization s.t. any associative edge $x \oplus y$ maps to an enode $(x = 0, y = 0)$, and for any repulsive edge pick either form. Hence in $N$ we have:

$\{e_i\}$ associative enodes from $S$, form $(s_i = 0, s_j = 0)$, $\{f_i\}$ associative enodes from $T$, form $(t_i = 0, t_j = 0)$, $\{a_i\}$ repulsive enodes $S \rightarrow T$, form $(s_i = 0, t_j = 1)$, $\{b_i\}$ repulsive enodes $S \leftarrow T$, form $(s_i = 1, t_j = 0)$.

Observe $N$ is bipartite with partitions $\{a_i, s_i^0, t_j^1, e_i\}$ and $\{b_i, s_i^1, t_j^0, f_i\}$. □

We now explore the case that $G$ has a frustrated cycle.

**Lemma 16.** *Any cycle $C$ in a binary pairwise MRF generates an induced (chordless) cycle $H$ in its NMRF $N$ with size at least as great, and with the same parity (odd/even number of vertices) as the number of repulsive edges (odd/even) in the MRF's cycle.*
*In particular, if $M$ contains any frustrated cycle with $\geq 4$ edges, or with 3 edges requiring any snode to link the enodes in $N$, then this maps to an odd hole in $N$.*

*Proof.* By Lemma 11, we may assume just one enode in $N$ per edge in $G$. Form a cycle $H$ in $N$ using the enodes corresponding to the edges of $C$, together with connecting snodes as required (if two enodes meet at a variable and have the same setting, add an snode with the opposite setting). Clearly $H$ is chordless and $|H| \geq |C|$.

Pick some enode $e_1$ and orientation around $H$. Consider the *end parity* of $e_1$, that is the setting for the next variable around $H$. For subsequent enodes, to maintain end parity requires an even (odd) total number of nodes, including possible snodes, for associative (repulsive) edges, and the reverse to flip end parity. Let $a_m$ and $a_f$ be the number of times end parity is maintained and flipped respectively using all associative edges around $H$, and similarly define $r_m$ and $r_f$ for all repulsive edges. In order to connect to the other end of $e_1$ after traversing $H$ requires in total (including $e_1$) an odd number of flips, hence $a_f + r_f \equiv 1 \pmod 2$. The total number of nodes in $H$ is comprised of the first enode together with all subsequent nodes, hence

$$|H| \equiv 1 + 0.a_m + 1.a_f + 1.r_m + 0.r_f \pmod 2$$
$$\equiv a_f + r_m + 1 \pmod 2 \equiv r_f + r_m \pmod 2. \square$$

Using Lemmas 13 and 16 we show the following result.

**Lemma 17.** *Let $M$ be a binary pairwise MRF that maps to an NMRF $N$. If $N$ is not perfect then $\exists$ a frustrated cycle in $M$ that maps to an odd hole in $N$. Hence, $N$ is perfect $\Leftrightarrow \not\exists$ such a cycle in $M$.*

*Proof.* By Lemma 13, $N$ contains an odd hole $H$. By Lemma 10, any snode in $H$ is adjacent to two enodes, and hence $H$ must have derived from a cycle in $M$. Lemma 16 completes the proof. □

**Lemma 18.** *The only 2-connected binary pairwise MRFs containing a frustrated cycle, that map to a perfect NMRF, are of the form $T_{m,n}$ or $U_n$.*

*Proof.* See section 2.2 for definitions. By Lemmas 16 and 17, we need only consider a frustrated triangle in $M$ whose enodes in $N$ require no connecting snodes. This triangle may have either (1) one repulsive and two associative edges, which we shall show must be of the form $U_n$ or $T_{m,n}$ with $n \geq 1$, or (2) three repulsive edges, which we shall show must be of the form $T_{m,n}$.

It is simple to check that, in either case, a fourth vertex adjacent to all 3 vertices of the triangle, resulting in a $K_4$ clique, does not admit a reparameterization that avoids a frustrated cycle requiring connecting snodes.

*Case 1: Triangle with one repulsive edge.* We have a $U_1$ structure. Let the configuration in the MRF be $s \oplus t \ominus v_1 \oplus s$. In order to avoid connecting snodes in $N$, we must have one of the following two reparameterizations: $\{(s=0, t=0), (t=1, v_1=0), (v_1=1, s=1)\}$ or $\{(s=1, t=1), (t=0, v_1=1), (v_1=0, s=0)\}$. Once one edge has been selected, the others can follow in only one way. Consider what may be added to this graph while remaining 2-connected and avoiding a frustrated cycle with $\geq 4$ edges. Any additional vertex $v_2$ must be attached by disjoint paths to at least 2 vertices $x$ and $y$ of the triangle. If either path has length $\geq 2$ then, by choosing one or other path in the original $U_1$ from $x$ to $y$, we always find a frustrated cycle with $\geq 4$ edges, leading to an odd hole. Using the argument from the preceding paragraph, $v_2$ must be adjacent to exactly 2 vertices of $U_1$. If these vertices are connected by an associative edge, we now have $U_2$; otherwise we have $T_{0,2}$. Checking cases now shows that the only way to add further vertices results in $U_n$ or $T_{m,n}$ structures, with any $m \geq 0, n \geq 1$ allowed.

*Case 2: Triangle with three repulsive edges.* We have $T_{1,0}$. Similar reasoning to case 1 shows that the only possibilities are $T_{m,n}$ for any $m \geq 1, n \geq 0$. □

Taking the results of this section together, we have the following characterization.

**Theorem 19.** *A binary pairwise MRF maps to a perfect NMRF for all valid $\psi_c$ iff each of its blocks (using all edges) has the form $B_R$, $T_{m,n}$ or $U_n$.*

## 5.1 REMARKS

Theorem 19 certainly has theoretical value in establishing the boundaries of the MWSS approach for this class of MRFs. Further, it broadens the landscape of tractable models. Each of the three block categories is itself tractable by other methods *in isolation*: QPBO (Rother et al., 2007) is guaranteed to be able to handle a $B_R$ structure (though not $T_{m,n}$ or $U_n$), or indeed a $B_R$ structure may be flipped to yield a fully associative model which can be solved with any appropriate technique such as graph cuts; and each $T_{m,n}$ or $U_n$ has low tree width so admits traditional inference methods. To our knowledge, however, our approach is the first to be able to handle an MRF containing $\Omega(n)$ of these structures, including high tree width $B_R$ sections, automatically in polynomial time.

### 5.1.1 Efficient Detection

Detecting if a binary pairwise MRF with topology $(V, E)$ satisfies our conditions may be performed in time $O(|E|)$: identifying block structure is an application of DFS, then each block type may be efficiently checked. The $T_{m,n}$ and $U_n$ structures are straightforward. For $B_R$, first test if it is bipartite using just $E_R$ (an application of BFS). Next check each component by $E_R$ to see that no $E_A$ cross partitions. Then stitch together partitions from different components (if more than one) using $E_A$. If any $E_A$ cross partitions then it is easy to see $\exists$ a frustrated cycle with $\geq 4$ edges which would lead to an odd hole in the NMRF.

## 6 HIGHER ORDER SUBMODULAR

As noted in the introduction, (Jebara, 2012) has shown that a fully associative binary pairwise model, which is equivalent to a model with supermodular pairwise $\psi$ functions (submodular cost functions), can always be reparameterized so as to yield a bipartite pruned NMRF. Indeed, we have seen in section 2.6 that, for each associative edge $x \oplus y$, one may reparameterize and prune the edge clique group so as to leave only either form $(x=0, y=0)$ or $(x=1, y=1)$. Here we extend the analysis to consider higher order models, still focusing on submodular cost functions over binary variables. We shall show that for potentials over 3 variables, a bipartite pruned NMRF is obtained for any topology iff all cost functions are submodular. Further, we show that submodularity is a necessary but strictly insufficient condition to obtain a bipartite pruned NMRF for all orders higher than 3.

Considering other approaches, this is similar to the result of (Zivny et al., 2009) that all order 3 submodular functions over Boolean variables can be represented by order 2 submodular functions using auxiliary variables, but this is not always true when the order $> 3$. Also, (Kolmogorov & Zabih, 2004) showed that submodularity was necessary for a function to be graph-representable. However, (Arora et al., 2012) recently demonstrated a novel graph cuts method for submodular cost functions of any order[8] over binary variables. Still, our result usefully clarifies the boundaries of our approach if we restrict to bipartite NMRFs only, and there is hope yet that a broader class of models may map to the wider class of perfect NMRFs.

---

[8] The time is exponential in the order of the potentials.

## 6.1 NOTATION

Let $\psi$ be an order $k$ potential function over $k$ binary variables $X = \{X_1, \ldots, X_k\}$. Let one setting be $x = (x_1, \ldots, x_k)$. Let $x - ij$ be a setting for all variables other than $X_i$ and $X_j$. Let $\psi_x = \psi(X = x)$. Define the supermodularity $s$ of $\psi$ wrt $X_i, X_j$ on the projection given by $x-ij$, as $s^{ij}_{x-ij} = \psi(X_i = 0, X_j = 0) + \psi(X_i = 1, X_j = 1) - \psi(X_i = 1, X_j = 0) - \psi(X_i = 0, X_j = 1)$ where all other variables in $X \setminus \{X_i, X_j\}$ are held fixed at $x - ij$.

Define $\alpha_k = \sum_{\text{all } 2^k \text{ settings of } x} (-1)^{\#0s \text{ in } x} \psi_x$. Observe that for $k = 2$, this is the supermodularity $s$ term. For $k = 3$, this is the difference between $s$ with (any) one variable set to 0 and that with the same variable set to 1. For $k = 4$, we have the sum of two $s$ terms minus two others, etc.

$\forall Y \subseteq \mathcal{P}(X)$, let $O_Y$ and $I_Y$ be weighted indicator functions. The $O$ functions are 0 unless all of $Y$ are 0. The $I$ functions are 0 unless all of $Y$ are 1. Otherwise, $O_Y$ and $I_Y$ take values $Z_Y$ and $A_Y$, respectively. $Y = b$ means fix variables $Y$ at value $b$ where $b \in \mathbb{B} = \{0, 1\}$.

In order to map to a bipartite pruned NMRF for any topology at order $k$, we must be able to represent every $\psi_x$ as the sum of a constant term and nonnegative[9] $O$ and $I$ indicator functions over all subsets of $X$, which correspond exactly to the nodes in the pruned NMRF (which is then clearly bipartite with stable sets corresponding to the $\{O_Y\}$ and $\{I_Y\}$).

## 6.2 RESULTS

**Theorem 20.** *For $k \geq 2$, mapping to a bipartite pruned NMRF for any topology $\Rightarrow \psi$ is supermodular, equivalently every projection of $\psi$ onto two variables is supermodular.*

*Proof.* Given the $\psi_x$ representation from the previous paragraph, consider which $A_Y, Z_Y$ terms survive when a general supermodularity term $s^{ij}_{x-ij}$ is computed. For some $Y$, analyze $A_Y$ terms (a similar result holds for $Z_Y$ terms): $Y$ will include either none, one or two of the variables $\{X_i, X_j\}$. Consider the cases: If none, then $A_Y$ does not feature in the $s^{ij}_{x-ij}$ computation. If one, then we get plus $A_Y$ (from the $X_i = X_j = 1$ term) minus $A_Y$ (from the appropriate other term), so they cancel. Finally, if two, then we simply get plus the $A_Y$ term. Hence for every $s^{ij}_{x-ij}$, it must be equal to the sum of some $A_{Y_i}$ and $Z_{Y_j}$ terms, all of which are constrained to be $\geq 0$. Hence all supermodularity terms are $\geq 0$. □

Further, for $k = 4$, it is easily checked that $\alpha_k = A_X + Z_X$, where we require $A_X, Z_X \geq 0$, yet it also equals $s^{ij}_{x-ij=00} + s^{ij}_{x-ij=11} - s^{ij}_{x-ij=01} - s^{ij}_{x-ij=10}$ (for any 2 variables $X_i, X_j$), which may be positive but equally may be negative.[10] Similarly for all $k > 4$, we are not able to represent all supermodular $\psi$ functions.

**Theorem 21.** *For general interactions over $k = 3$ variables, $\psi$ is supermodular $\Leftrightarrow$ we obtain a bipartite pruned NMRF for any topology.*

*Proof.* ($\Leftarrow$) follows from Theorem 20. ($\Rightarrow$) we provide a constructive proof:[11]

If $\alpha_k \geq 0$, use only $O_Y$ for $|Y| \geq 2$. Set $Z_X = \alpha_k$. For $|Y| = 2$, set $Z_Y = s^Y_1$. For $|Y| = 1$, set $Z_Y = \psi(Y = 0, (X \setminus Y) = 1) - \psi_{111}$. Set constant to $\psi_{111}$ to observe we match $\psi_x$ values $\forall x$. Now reparameterize all singleton terms and prune as usual, see section 2.5.

If $\alpha_k \leq 0$, use only $I_Y$ for $|Y| \geq 2$. Set $A_X = -\alpha_k$. For $|Y| = 2$, set $A_Y = s^Y_0$. For $|Y| = 1$, set $A_Y = \psi(Y = 1, (X \setminus Y) = 0) - \psi_{000}$. As before, set constant to $\psi_{000}$ to check values, then reparameterize all singleton terms and prune, see section 2.5. □

## 7 CONCLUSIONS

The MWSS approach to MAP inference is an exciting, recent approach, leveraging the rapid progress in combinatorics. Here we have derived new general tools (section 3), defined the scope of the approach in an important, broad setting (sections 4 and 5), where we were able to extend the range of known tractable models, and clarified the power of mapping to bipartite NMRFs (section 6).

Future areas to explore include non-bipartite perfect NMRFs for higher order potentials, and variables with a greater number of labels.

**Acknowledgments**

We thank Maria Chudnovsky for helpful discussions and for the shorter proof of Theorem 7 shown, and the anonymous reviewers for valuable suggestions. This material is based upon work supported by the National Science Foundation under Grant No. 1117631.

---

[9] It is critical that the functions be nonnegative in order that the corresponding nodes in the NMRF are the only ones not pruned.

[10] An example of supermodular $\psi$ for $k = 4$ where $\alpha_k < 0$: $\psi(x_1, x_2, x_3, x_4) = 0$ except $\psi(0,0,0,0) = 2, \psi(1,0,0,0) = \psi(0,1,0,0) = \psi(0,0,1,0) = \psi(0,0,0,1) = 1$.

[11] In fact, as shown, we need use only either exclusively $O_Y$ or $I_Y$ nodes for $|Y| \geq 2$, which may further improve efficiency.